\documentclass[conference]{IEEEtran}
\usepackage[utf8]{inputenc}
\usepackage{array}
\usepackage{wrapfig}
\usepackage{multirow}
\usepackage{tabu}
\usepackage{xcolor}
\usepackage{cite}
\usepackage{amsmath,amssymb,amsfonts}
\usepackage{algorithmic}
\usepackage{algorithm}
\usepackage{graphicx}
\usepackage{textcomp}
\usepackage{caption}
\usepackage{comment}
\usepackage{subcaption}
\usepackage{float}
\usepackage{eucal}
\usepackage{soul}
\usepackage{fixltx2e}
\usepackage{enumerate}
\usepackage{cuted}
\usepackage{tabularx}
\usepackage{pgf-pie}
\usepackage{enumitem}
\usepackage{flushend}
\usepackage{booktabs}
\def\BibTeX{{\rm B\kern-.05em{\sc i\kern-.025em b}\kern-.08em
    T\kern-.1667em\lower.7ex\hbox{E}\kern-.125emX}}
\usepackage{fancyhdr}

\fancypagestyle{plain}{
  \fancyhf{}
  \fancyhead[C]{}     
  \fancyfoot[L]{}

}
\usepackage{eso-pic}

\begin{document}
\title{SPARC: Subspace-Aware Prompt Adaptation for Robust Continual Learning in LLMs}
\author{
\IEEEauthorblockN{Dinithi Jayasuriya\IEEEauthorrefmark{1}, 
Sina Tayebati\IEEEauthorrefmark{1}, 
Davide Ettori\IEEEauthorrefmark{1}, 
Ranganath Krishnan\IEEEauthorrefmark{2}, 
Amit Ranjan Trivedi\IEEEauthorrefmark{1}}
\IEEEauthorblockA{\IEEEauthorrefmark{1}Department of Electrical and Computer Engineering, University of Illinois Chicago, Chicago, IL \\
Email: \{dkasth2, stayeb3, detto3, amitrt\}@uic.edu}
\IEEEauthorblockA{\IEEEauthorrefmark{2}Intel Labs, Oregon \\
Email: ranganath.krishnan@intel.com}
}
\maketitle
\begin{abstract}
We propose SPARC, a lightweight continual learning framework for large language models (LLMs) that enables efficient task adaptation through prompt tuning in a lower-dimensional space. By leveraging principal component analysis (PCA), we identify a compact subspace of the training data. Optimizing prompts in this lower-dimensional space enhances training efficiency, as it focuses updates on the most relevant features while reducing computational overhead. Furthermore, since the model’s internal structure remains unaltered, the extensive knowledge gained from pretraining is fully preserved, ensuring that previously learned information is not compromised during adaptation. Our method achieves high knowledge retention in both task-incremental and domain-incremental continual learning setups while fine-tuning only 0.04\%\ of the model’s parameters. Additionally, by integrating LoRA, we enhance adaptability to computational constraints, allowing for a tradeoff between accuracy and training cost. Experiments on the SuperGLUE benchmark demonstrate that our PCA-based prompt tuning combined with LoRA maintains full knowledge retention while improving accuracy, utilizing only 1\%\ of the model’s parameters. These results establish our approach as a scalable and resource-efficient solution for continual learning in LLMs.
\end{abstract}

\section{Introduction}
Large Language Models (LLMs) have demonstrated remarkable capabilities in natural language processing, enabling transfer learning across diverse tasks and domains. Their ability to encode rich semantic representations allows them to generalize effectively and adapt to new tasks with minimal supervision. These characteristics, along with their proficiency in few-shot learning, have established LLMs as indispensable tools for applications such as text generation~\cite{kasner2024beyond}, question answering~\cite{saito2024unsupervised}, summarization~\cite{mullick2024leveraging}, and reasoning~\cite{huang2022towards}.

While pretrained LLMs excel in handling static datasets and fixed tasks, many real-world applications require dynamic adaptability to evolving environments. For instance, LLM-based autonomous navigation systems must respond to unseen driving conditions, terrains, and tasks, such as understanding new traffic laws or adapting to varying weather conditions. Retraining a new model for each task is computationally prohibitive, especially for large-scale systems, and the assumption that data from earlier tasks is always available for retraining is often unrealistic due to storage limitations, regulatory constraints, or privacy concerns. These challenges necessitate methods that enable models to integrate new information incrementally while retaining the prior knowledge.

However, LLMs also face critical challenges in \textit{continual learning}--the ability to acquire new knowledge without overwriting previously learned information. Naive fine-tuning on new tasks often leads to \textit{catastrophic forgetting} \cite{luo2023empirical}, where updates disrupt prior knowledge. The high computational cost of fine-tuning all parameters for each task is impractical, given that most of the model's parameters already encode transferable knowledge. Thus, continual learning in LLMs requires approaches that can facilitate efficient task-specific adaptation while preserving general-purpose capabilities.

Several strategies have been proposed to address these challenges: \textit{Replay-based methods} \cite{bagus2021investigation} mitigate forgetting by replaying data from previous tasks during training. However, these methods incur significant memory overhead and may be unsuitable for sensitive domains where storing and replaying data raises privacy concerns \cite{de2021continual}. \textit{Regularization-based approaches}, such as Elastic Weight Consolidation (EWC) \cite{kirkpatrick2017overcoming} and Synaptic Intelligence (SI) \cite{zenke2017continual}, attempt to preserve knowledge by penalizing updates to parameters critical to previous tasks. However, these methods are less effective in the high-dimensional parameter spaces of LLMs. \textit{Parameter-efficient fine-tuning (PEFT)} methods, including LoRA \cite{hu2021lora,dou2023loramoe,liang2024inflora}, and adapter tuning \cite{gao2023unified}, address scalability and efficiency by introducing lightweight task-specific modules that reduce trainable parameters. While PEFT methods are computationally efficient, they often require architectural modifications, complicating deployment and risking interference with pretrained representations.

Prompt tuning has emerged as an alternative that is well-suited for task-specific adaptation in LLMs. By introducing small, trainable embeddings (soft prompts) appended to input tokens, prompt tuning enables task-specific learning while keeping the base model frozen. This approach significantly reduces the number of trainable parameters, making it computationally efficient and scalable for large models. However, existing prompt tuning methods typically train a separate prompt for each task, resulting in inefficiencies and redundancy when tasks share underlying representations. Moreover, these methods lack mechanisms to address catastrophic forgetting, limiting their applicability to continual learning scenarios where sequential task adaptation is required.

To address these limitations, as in Fig. 1(a), we propose SPARC, a novel framework for continual learning in LLMs that leverages \textit{subspace-guided prompt tuning}. At its core, the framework represents each task as a subspace in the input embedding space using principal component analysis (PCA). By quantifying subspace overlap, the framework determines whether a new task can reuse an existing prompt or requires a new one. This reduces computational overhead by reusing prompts for tasks with overlapping representations, thereby facilitating knowledge transfer across related tasks. For tasks with minimal overlap, new prompts are initialized in orthogonal subspaces to preserve task-specific independence and mitigate interference. This alignment prevents catastrophic forgetting by ensuring that new knowledge does not overwrite previously learned representations.

A key advantage of the proposed framework is its parameter efficiency. By limiting updates to soft prompts, the framework minimizes the number of trainable parameters without requiring architectural modifications to the base model, ensuring scalability to large LLMs. This distinguishes it from approaches like LoRA and adapter tuning, which require changes to the model's structure and may disrupt its pretrained capabilities. The framework\textquotesingle{}s architecture-agnostic design and lightweight computational requirements make it particularly suitable for real-world applications where resource efficiency and model integrity are critical.

We validate the framework through extensive experiments in both \textit{domain-incremental learning}, where the model adapts across datasets from distinct domains, and \textit{task-incremental learning}, where it sequentially learns tasks with different objectives. Our results demonstrate that the framework effectively mitigates catastrophic forgetting while achieving strong forward transfer, enabling efficient learning of new tasks, and backward transfer, ensuring a 97\%\ prior knowledge retention. In \textit{domain-incremental learning}, the framework maintains an average forgetting ratio of 3\%, while in \textit{task-incremental learning}, it achieves no forgetting, demonstrating its robustness in continual learning scenarios. Notably, these results are obtained while fine-tuning only 0.04\% of the model’s parameters, significantly reducing computational overhead compared to baseline approaches. By combining \textit{prompt reusability} and \textit{orthogonal subspace alignment}, our method provides a scalable and resource-efficient solution for continual learning in LLMs, ensuring both adaptability and knowledge retention.

\section{Background}

\subsection{Soft Prompts and Soft Embeddings}
Prompt tuning is a parameter-efficient approach for adapting frozen pretrained LLMs to specific tasks without modifying their internal weights. It introduces trainable embeddings, called \textit{soft prompts}, which are prepended to the input token embeddings and optimized while keeping the LLM parameters fixed~\cite{lester2021power}. These learnable vectors, with dimensions matching the model’s embedding space, act as lightweight adapters to guide the model’s behaviour toward task-specific objectives while preserving its general-purpose knowledge.

The process begins with the initialization of soft prompts, which can be random or informed by task-specific heuristics. These prompts are concatenated with the input token embeddings and passed through the frozen LLM. During training, only the soft prompts are updated using task-specific gradients, while the pretrained weights remain unchanged. This selective optimization significantly reduces computational overhead. For instance, the trainable parameters are proportional to the product of the number of soft prompt tokens (\(T\)) and the embedding dimensionality (\(d\)), typically amounting to a few thousand parameters for large-scale models.

Soft prompts are particularly advantageous for large LLMs as they enable task-specific adaptation without requiring extensive computational resources or architectural modifications. Moreover, task-specific prompts can be stored independently and reused across tasks, making this approach ideal for multi-task learning and frequent updates. However, conventional prompt tuning methods lack mechanisms to mitigate \textit{catastrophic forgetting}, where updates for new tasks overwrite knowledge from earlier ones. Additionally, they often train redundant prompts for tasks with overlapping representations, limiting efficiency in continual learning scenarios.

\subsection{PCA-based Subspace Identification}
Principal Component Analysis (PCA) is a dimensionality reduction technique that transforms high-dimensional data into a smaller set of orthogonal axes, called principal components, which capture the maximum variance in the data~\cite{abdi2010principal}. By projecting data onto these components, PCA identifies the most significant patterns while preserving as much information as possible, making it particularly useful for analyzing and comparing embedding spaces across tasks.

Given a dataset \( \mathbf{X} \in \mathbb{R}^{N \times D} \), where \( N \) is the number of samples and \( D \) is the feature dimension, PCA projects \( \mathbf{X} \) into a lower-dimensional subspace \( \mathbb{R}^k \), where \( k < D \). The process begins by centering the data as \( \mathbf{X}_c = \mathbf{X} - \mu \), where \( \mu = \frac{1}{N} \sum_{i=1}^N \mathbf{x}_i \). Next, the covariance matrix \( \mathbf{C} = \frac{1}{N} \mathbf{X}_c^\top \mathbf{X}_c \) is computed, and eigenvalue decomposition is performed: \( \mathbf{C} \mathbf{w}_i = \lambda_i \mathbf{w}_i \), where \( \mathbf{w}_i \) are eigenvectors (principal components) and \( \lambda_i \) are eigenvalues. The top \( k \) eigenvectors corresponding to the largest eigenvalues are selected to form the transformation matrix \( \mathbf{W} \in \mathbb{R}^{k \times D} \), ensuring orthogonality (\( \mathbf{W} \mathbf{W}^\top = \mathbf{I} \), where \( \mathbf{I} \) is the identity matrix). 

These components define a task’s subspace in the embedding space. The projection of the original data into this subspace is given by \( \mathbf{X}_k = \mathbf{X}_c \mathbf{W}^\top \). PCA is particularly effective for continual learning as it enables tasks to be represented in lower-dimensional subspaces that capture their dominant features. These subspaces can then be compared to assess similarity across tasks, facilitating prompt reuse for tasks with shared representations or initializing orthogonal subspaces for independent tasks. By focusing on the most informative components, PCA reduces dimensionality while maintaining the core structure of task-specific knowledge.

 \begin{figure*}
     \centering
     \includegraphics[width=\linewidth]{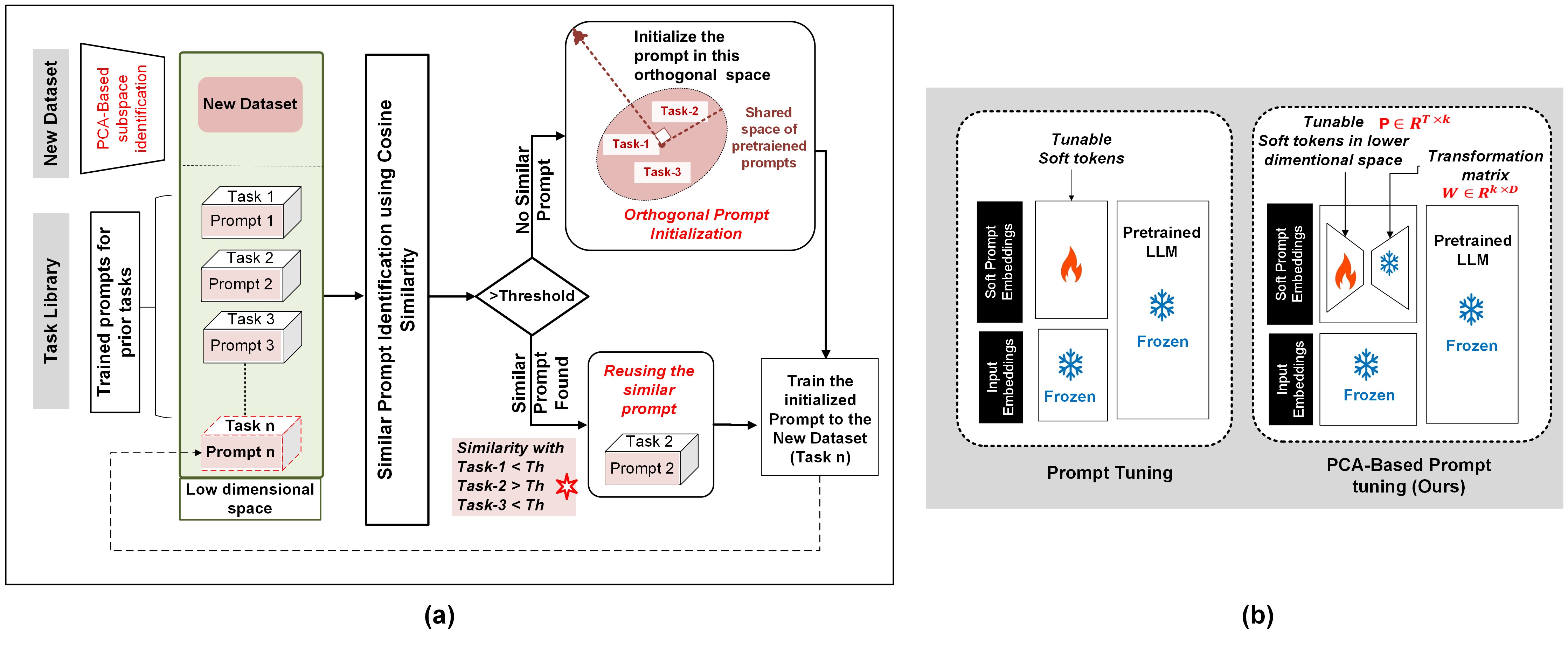}
     \caption{\textbf{Overview of SPARC:} \textbf{(a)} The subspace of the new dataset is computed using PCA. By measuring the cosine similarity between this new subspace and previously learned prompt subspaces, the framework determines whether a similar prompt already exists. If a match is found, the existing prompt is reused for initialization, enhancing knowledge retention. Otherwise, a new prompt is initialized in an orthogonal subspace to the existing prompts, ensuring differentiation and efficient adaptation. \textbf{(b)} The prompt embeddings consist of two key components: tunable soft tokens and a PCA-based transformation matrix. This design significantly reduces the number of trainable parameters compared to traditional prompt tuning methods, making the approach more efficient while preserving model adaptability.}
 \end{figure*}
 
\subsection{Subspace Overlap using Cosine Similarity}
To enable efficient knowledge transfer across tasks, subspace overlap is quantified using cosine similarity\cite{passalis2018unsupervised}. This technique measures the alignment between principal components of embedding subspaces from different tasks, helping determine whether knowledge from a previous task can be reused for a new one. For two datasets \( D_1 \) and \( D_2 \), with principal components \( P_1 = \{\mathbf{p}_1^1, \dots, \mathbf{p}_k^1\} \) and \( P_2 = \{\mathbf{p}_1^2, \dots, \mathbf{p}_k^2\} \), the cosine similarity between components \( \mathbf{p}_i^1 \in P_1 \) and \( \mathbf{p}_j^2 \in P_2 \) is computed as:
\begin{equation}
\mathbf{S}_{ij} = \frac{\mathbf{p}_i^1 \cdot \mathbf{p}_j^2}{\|\mathbf{p}_i^1\| \|\mathbf{p}_j^2\|}.
\end{equation}
A principal component \( \mathbf{p}_i^1 \) is considered aligned with \( P_2 \) if its similarity with any \( \mathbf{p}_j^2 \) exceeds a predefined threshold \( \tau \). This alignment is summarized as the \textit{overlap percentage}, which quantifies how many components of \( P_1 \) match with \( P_2 \):
\begin{equation}
\text{Overlap Percentage} = \frac{\text{Number of Aligned Components}}{\text{Total Components in } P_1}.
\end{equation}

By identifying subspace overlap, the framework determines whether a prompt for a new task can reuse an existing one, minimizing redundancy. For tasks with substantial overlap, the corresponding prompt can be reused with minimal fine-tuning. For tasks with minimal overlap, a new prompt is initialized to capture the independent features of the new task. This enables the framework to reuse learned knowledge for related tasks while isolating representations for unique task characteristics.

\subsection{Orthogonal Subspaces for Task Separation}
Orthogonal subspaces are employed to prevent interference between tasks, ensuring that task-specific knowledge remains independent \cite{liang2024inflora}. This is particularly important for tasks with minimal overlap in their embedding subspaces, where reusing a previously learned prompt may lead to performance degradation or \textit{catastrophic forgetting}.

Two subspaces, \( P_{t+1} \) (new task) and \( P_i \) (previous task), are orthogonal if the inner product of all basis vectors between them is zero. Mathematically, this is expressed as:
\[
\langle P_{t+1}, P_i \rangle = 0, \quad \forall i = 1, 2, \dots, t.
\]
In practice, orthogonality is enforced by projecting the new subspace \( P_{t+1} \) onto the complement of the span of the previously learned subspaces \( \{P_1, P_2, \dots, P_t\} \). This ensures:
\[
\text{Proj}_{P_i}(\mathbf{u}) = 0, \quad \forall~\mathbf{u} \in P_{t+1}, \; P_i \in \{P_1, P_2, \dots, P_t\}.
\]

Orthogonal subspaces allow the framework to isolate new task-specific knowledge while preserving prior representations. By aligning new prompts with orthogonal directions in the embedding space, the model effectively mitigates catastrophic forgetting and interference \cite{chang2005orthogonal}. This approach also ensures that existing prompts remain intact and reusable for earlier tasks. Combining subspace overlap analysis with orthogonal initialization allows the framework to balance knowledge retention and efficient adaptation. Overlapping subspaces enable prompt reuse for related tasks, while orthogonal subspaces ensure task-specific independence for novel features, making the framework robust for continual learning of LLMs.

\section{SPARC: Subspace-Aware Prompt Tuning for Continual Learning}
The proposed framework leverages prompt tuning with subspace-guided initialization and orthogonal alignment to enable efficient and scalable continual learning of LLMs. Soft prompt embeddings are appended to the input tokens of a pretrained language model, allowing task-specific adaptation without modifying the model's internal weights. This section outlines the processes of prompt initialization, subspace overlap analysis, and orthogonal prompt generation to effectively balance knowledge transfer and task-specific independence.

\subsection{Prompt Initialization and Subspace Representation}
To adapt LLMs to new tasks, trainable soft prompts are initialized and optimized to represent the core characteristics of a dataset. The initialization process relies on PCA to capture the dominant features of the dataset’s embedding space. These principal components are leveraged for prompt initialization for parameter efficiency and task-specific adaptation (Fig. 1(b)).

\subsubsection{Embedding Principal Components via PCA}
Given a dataset \( d_i \), its embeddings \( \mathbf{X} \in \mathbb{R}^{N \times D} \) are generated by passing input tokens through a pretrained LLM or a sentence transformer, where \( N \) is the number of samples, and \( D \) is the embedding dimension. PCA is applied to \( \mathbf{X} \) to extract the top \( k \) components forming a reduced subspace \( \mathbb{R}^k \), where \( k \ll D \).

The transformation matrix \( \mathbf{W} \in \mathbb{R}^{k \times D} \) is computed using the eigenvectors of the covariance matrix:
\begin{equation}
\mathbf{C} = \frac{1}{N} \mathbf{X}_c^\top \mathbf{X}_c, \quad \text{where } \mathbf{X}_c = \mathbf{X} - \mu, \; \mu = \frac{1}{N} \sum_{i=1}^N \mathbf{x}_i.
\end{equation}

The top \( k \) eigenvectors of \( \mathbf{C} \) define the principal components in \( \mathbf{W} \), capturing the directions of maximum variance. Embeddings are projected into the subspace as \( \mathbf{X}_k = \mathbf{X}_c \mathbf{W}^\top \), retaining the most informative features of the dataset while reducing noise and dimensionality. This subspace forms the initialization space for task-specific prompts.

\subsubsection{Learnable Prompt Parameters}
Soft prompts are initialized as a trainable parameter matrix \( \mathbf{P} \in \mathbb{R}^{T \times K} \), where \( T \) is the number of soft tokens, and \( K \) is the dimensionality of the principal subspace (\( k \)). Each row of \( \mathbf{P} \) corresponds to a trainable embedding vector, which is prepended to the input token embeddings of \( d_i \). During training, only \( \mathbf{P} \) is updated to minimize the task-specific loss, while the pretrained model’s weights remain frozen. This selective optimization enables efficient task adaptation with minimal computational overhead. For example, with \( T = 10 \) and \( K = 100 \), the number of trainable parameters is reduced to 1,000, which is orders of magnitude smaller than fine-tuning the entire LLM. By decoupling prompt updates from the model, the framework ensures modularity and scalability across tasks.

\subsection{Subspace Overlap Analysis}
The framework determines whether a new task can reuse an existing prompt or requires a new one by evaluating the similarity between the subspace of the new dataset \( d_{t+1} \) and those of previously encountered datasets \( \{d_1, d_2, \dots, d_t\} \).

\subsubsection{Quantifying Overlap Using Cosine Similarity}
Let \( P_{t+1} = \{\mathbf{p}_1^{t+1}, \dots, \mathbf{p}_k^{t+1}\} \) represent the top \( k \) principal components of \( d_{t+1} \), and \( P_i \) the components of an existing dataset \( d_i \). Cosine similarity is used to measure alignment between the principal components:
\begin{equation}
\mathbf{S}_{jk} = \frac{\mathbf{p}_j^{t+1} \cdot \mathbf{p}_k^i}{\|\mathbf{p}_j^{t+1}\| \|\mathbf{p}_k^i\|}.
\end{equation}
For each \( \mathbf{p}_j^{t+1} \), the maximum similarity score across all components in \( P_i \) is recorded. The \textit{overlap percentage} quantifies the proportion of components in \( P_{t+1} \) that align with \( P_i \) above a similarity threshold \( \tau \):
\begin{equation}
\text{Overlap Percentage} = \frac{\text{Number of Aligned Components}}{\text{Total Components in } P_{t+1}}.
\end{equation}

If the overlap percentage exceeds \( \tau \) (e.g., 50\%), the corresponding prompt \( p_i \) is reused with minimal fine-tuning. For tasks with low overlap, a new prompt is initialized to capture the unique characteristics of \( d_{t+1} \).

\subsection{Orthogonal Prompt Initialization}
For datasets with minimal subspace overlap, the framework initializes a new prompt in an orthogonal subspace to ensure task-specific independence and prevent interference with previously learned prompts.

\subsubsection{Orthogonal Projection}
Given the subspaces \( \{P_1, P_2, \dots, P_t\} \) of previously learned tasks, the orthonormal basis \( \mathbf{V} \) is constructed by concatenating their principal components. The embeddings of \( d_{t+1} \), denoted \( \mathbf{X}_{t+1} \), are projected onto \( \mathbf{V} \) to compute the aligned components:
\begin{equation}
\text{Proj}_{\mathbf{V}}(\mathbf{X}_{t+1}) = \mathbf{X}_{t+1} \mathbf{V} \mathbf{V}^\top.
\end{equation}
The orthogonal component is derived as:
\begin{equation}
\mathbf{X}_{\text{orth}} = \mathbf{X}_{t+1} - \text{Proj}_{\mathbf{V}}(\mathbf{X}_{t+1}).
\end{equation}

PCA is then applied to \( \mathbf{X}_{\text{orth}} \) to identify its principal components, forming a subspace that captures the unique features of \( d_{t+1} \). The new prompt is initialized in this orthogonal subspace to ensure independence from existing tasks.

\subsubsection{Training and Adaptation}
The orthogonally initialized prompt \( \mathbf{P}_{t+1} \) is optimized to minimize the task-specific loss. Existing prompts remain frozen and accessible for inference, allowing the framework to leverage previously acquired knowledge while integrating new information. This ensures robust continual learning with strong forward and backward transfer. By combining subspace overlap analysis and orthogonal initialization, the framework balances knowledge reuse for related tasks and independence for novel features, addressing the core challenges of catastrophic forgetting and task-specific adaptation in LLMs.

\begin{figure*}
    \centering
    \includegraphics[width=0.85\linewidth]{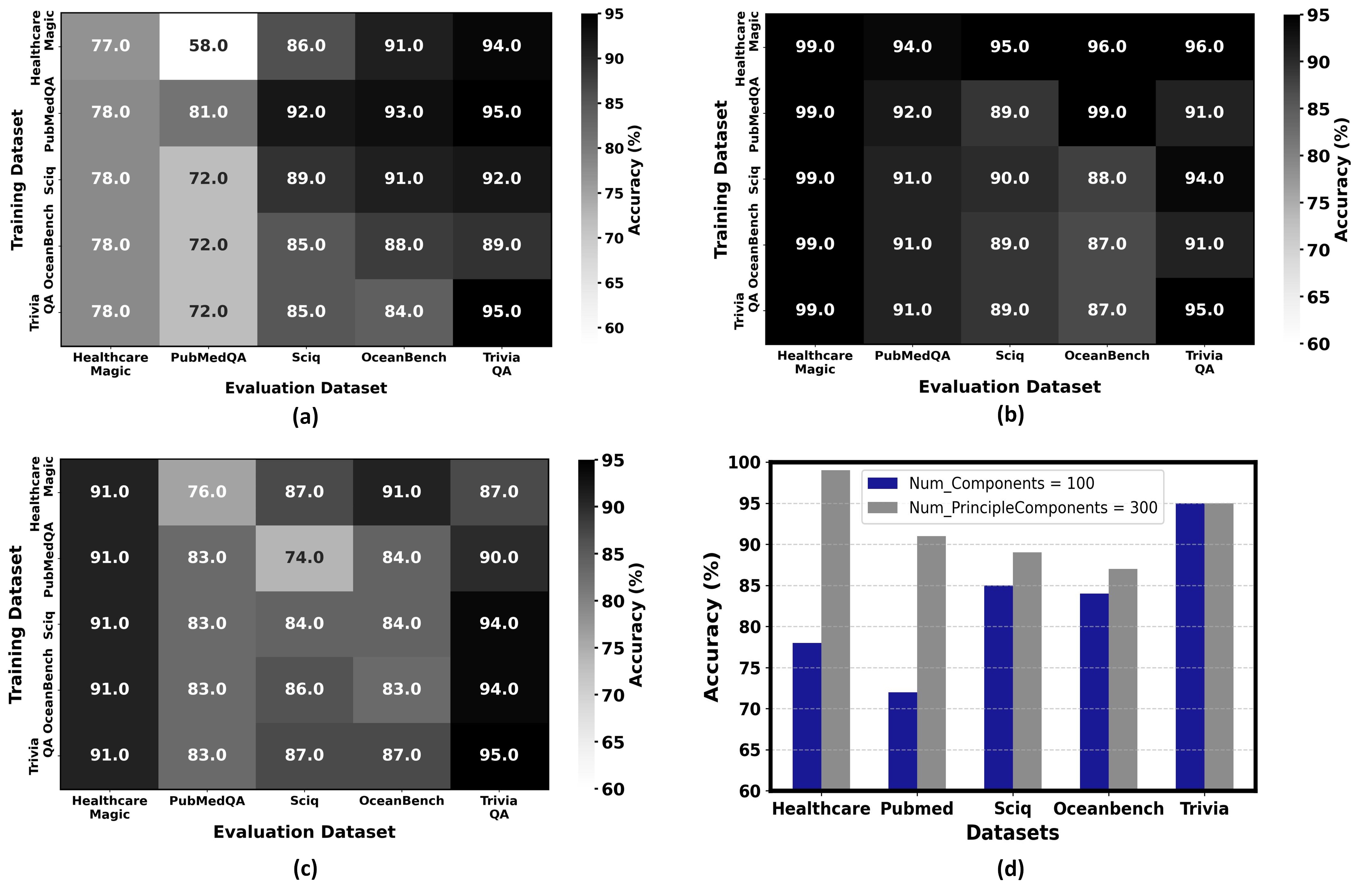}
    \caption{\textbf{Per-token accuracy with varying numbers of soft tokens and principal components:}  
    (a) 100 principal components, 20 soft tokens.  
    (b) 300 principal components, 20 soft tokens.  
    (c) 100 principal components, 40 soft tokens. (d) Accuracy with different number of PCA components }
    \label{fig:heatmaps}
\end{figure*}

\section{Results}

\subsection{Evaluation Setup}
We evaluated our framework in \textit{domain-incremental} and \textit{task-incremental} learning settings using GPT-2 and DeBERTa-base as base models. For domain-incremental learning, we use five datasets spanning healthcare, scientific literature, and general knowledge, sequentially introducing them to simulate real-world incremental learning. Per-token accuracy measures the framework’s ability to generate outputs aligned with task-specific prompts while retaining previously learned knowledge. For task-incremental learning, accuracy is determined by selecting the most likely class from model logits and assessing the model’s ability to classify tasks incrementally.

\subsection{Impact of Principal Components and Soft Tokens}
To analyze the effect of \textit{principal components} in PCA-based subspace selection, we conducted experiments with 100 and 300 components. The results in Fig. 2(d) show that increasing components improves accuracy for domains like healthcare and scientific literature, where richer representations are beneficial. However, for general knowledge tasks, performance remains stable, indicating that fewer components suffice. This underscores the importance of selecting an optimal number of PCA components to balance accuracy and computational efficiency. We also examine the impact of \textit{soft tokens}. While increasing tokens enhances task-specific performance by introducing more trainable parameters, it slightly reduces forward transfer due to overfitting to domain-specific features. These findings highlight the trade-off between specialization and generalization in continual learning.

\subsection{Domain-Incremental Learning Performance}

\begin{figure}[t!]
    \centering
    \includegraphics[width=0.48\linewidth,height=0.479\linewidth]{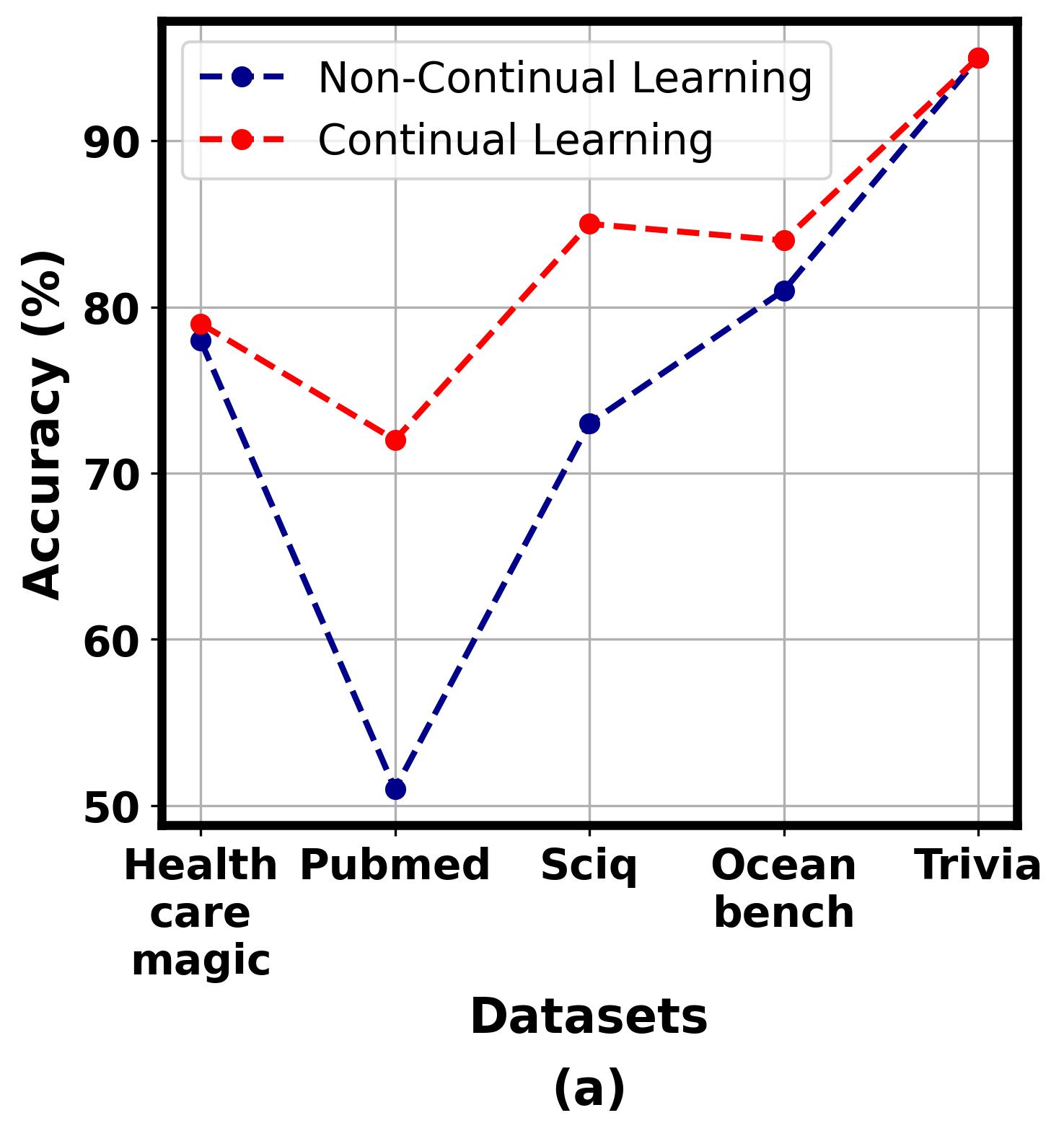}
    \includegraphics[width=0.48\linewidth,height=0.479\linewidth]{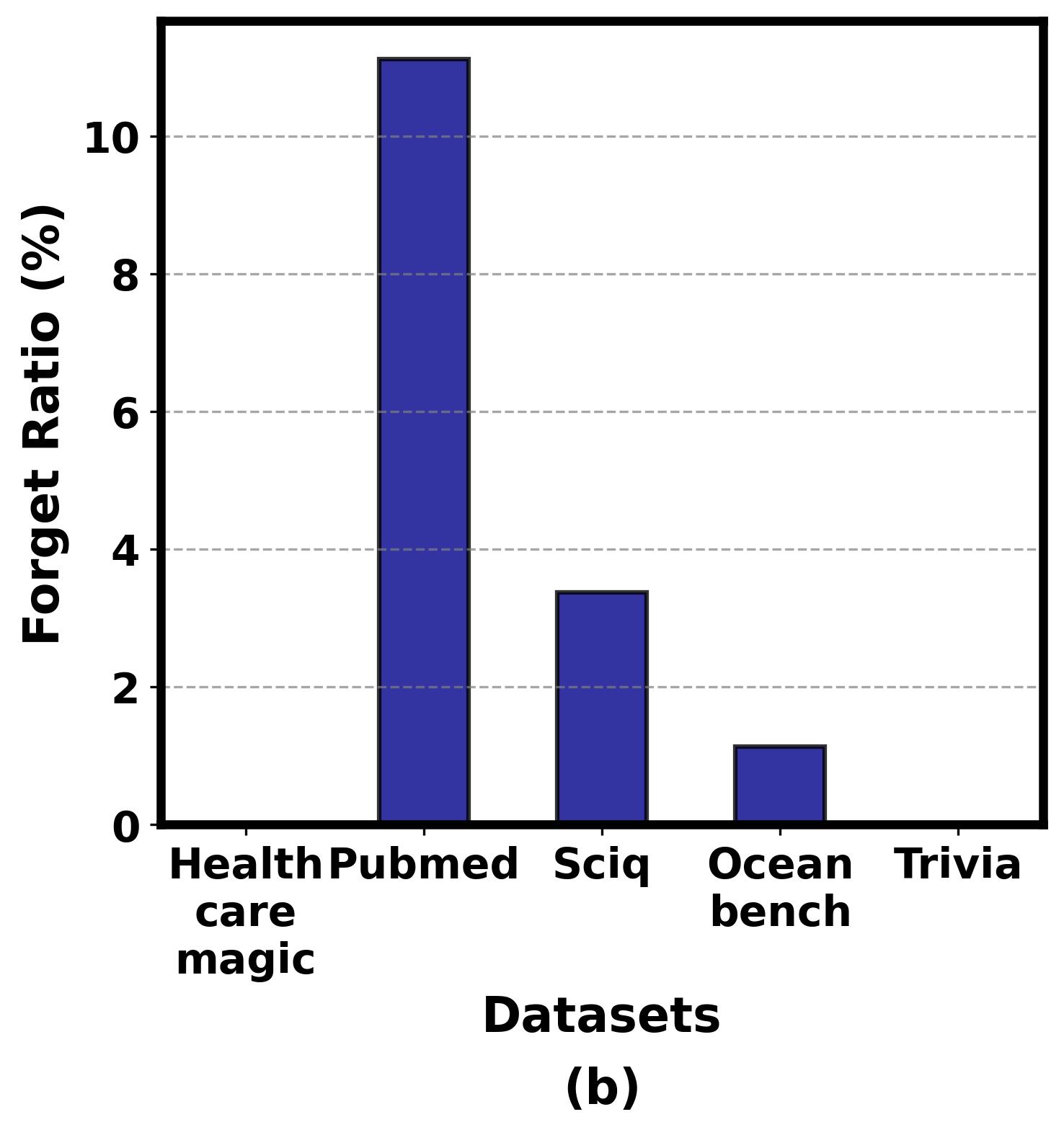}
    \caption{\textbf{Accuracy in Domain Incremental Learning.} \\ (a) Non-Continual learning Accuracy is calculated by fine-tuning the model individually on each dataset, while final accuracy is obtained by sequentially fine-tuning the model across datasets. (b) Forgetting ratio for each dataset after the completion of sequential training.}
\label{fig:forget_ratio_DL}
\end{figure}

The domain-incremental evaluation includes datasets from diverse fields. HealthcareMagic and PubMedQA~\cite{jin2019pubmedqa} cover medical question-answering, requiring models to infer responses based on patient queries and biomedical literature. SciQ~\cite{SciQ} focuses on scientific reasoning, testing the model’s ability to answer science-related multiple-choice questions. OceanBench~\cite{bi2023oceangpt} introduces real-world oceanographic data analysis, challenging the model’s adaptability to highly specialized knowledge and TriviaQA~\cite{joshi2017triviaqa} evaluates general knowledge comprehension by requiring the model to answer fact-based questions across a broad range of topics. This diverse dataset selection ensures a rigorous assessment of the framework’s capacity to adapt across distinct domains while preserving prior knowledge. The sequential training was conducted in the order of the x-axis in Fig. 2(a). We evaluated our framework on forward and backward transfers. Forward transfer in continual learning refers to the model’s ability to leverage previously learned knowledge to improve performance on new tasks without additional retraining \cite{parisi2019continual}. Conversely, backward transfer assesses whether learning a new task enhances or preserves performance on previously encountered tasks, ensuring knowledge retention and mitigating catastrophic forgetting. The results in Fig. 2  exhibit strong forward transfer, as demonstrated by improved accuracy on unseen datasets immediately after training on a new domain. For example, fine-tuning on a healthcare dataset significantly enhances performance on scientific literature datasets, leveraging shared semantic structures via subspace-guided prompt initialization. Backward transfer is also robust, with minimal accuracy degradation on earlier datasets after sequential training. Fig. 3(b) shows the forgetting ratio, measured as the percentage drop in accuracy on prior datasets, consistently remains below 5\%. This highlights the framework’s effectiveness in mitigating catastrophic forgetting.

We compare our framework's final accuracy after completing sequential training with the baseline accuracy obtained by fine-tuning the model individually on each dataset. Fig. 3(a) shows that our approach consistently outperforms baseline fine-tuning, demonstrating superior knowledge retention and transfer. For instance, after training on healthcare, scientific, and general knowledge datasets, final accuracy exceeds the baseline by a significant margin. These results validate the efficiency of subspace-guided prompt reuse and orthogonal prompt initialization in balancing task-specific learning and generalization. 

\subsection{Task-Incremental Learning Performance}

\begin{figure}[t!]
    \centering
    \includegraphics[width=1\linewidth]{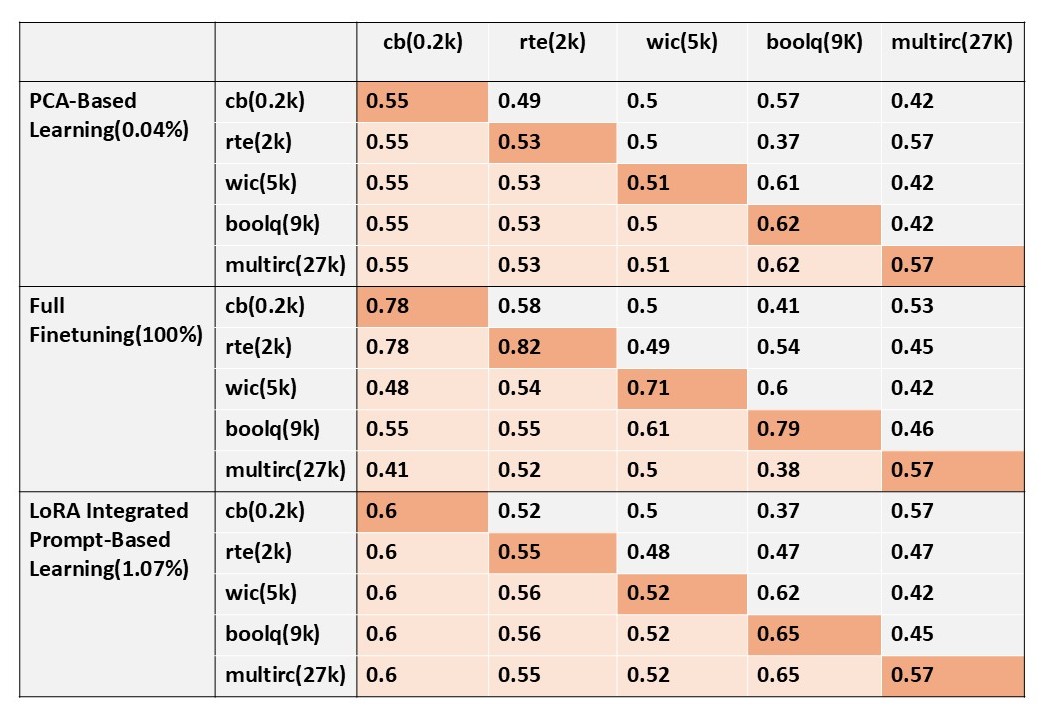}
    \caption{\textbf{Performance Comparison of Task-Incremental
Learning Methods:} Accuracy results for three approaches, PCA-Based Learning,
Full Finetuning, and LoRA-Integrated Prompt-Based Learning.}
\label{fig:table}
\end{figure}

The framework was also evaluated on diverse NLP tasks, using five SuperGLUE benchmark datasets such as CB (CommitmentBank), RTE (Recognizing Textual Entailment), WiC (Word-in-Context), BoolQ (Boolean Question Answering), and MultiRC (Multi-Sentence Reading Comprehension) \cite{wang2019superglue}. These datasets cover a diverse range of NLP tasks including reasoning over short texts, contextual word sense disambiguation, answering yes/no questions based on unstructured text, and extracting relevant information from multi-sentence passages. This diversity ensures a comprehensive evaluation of the framework’s ability to generalize across different linguistic challenges in a continual learning setting. Sequential training is performed to assess both forward transfer (efficient adaptation to new tasks) and backward transfer (retention of previously learned knowledge). Results confirm the framework’s flexibility and robustness across varying task objectives. Fig.~\ref{fig:table} presents evaluation results for \textit{Full Fine-Tuning, PCA-based Continual Learning}, and \textit{LoRA-integrated Continual Learning}. During training, datasets are introduced sequentially, with accuracy measured on both previously trained and unseen datasets after each step. The diagonal values indicate accuracy on the current dataset, while other values represent performance on past and unseen datasets. \textit{Full Fine-Tuning} achieves the highest accuracy but requires substantial computational resources, fine-tuning 100\% of model parameters, making it impractical for scalable continual learning. In contrast, our PCA-based method achieves competitive accuracy while fine-tuning only 0.04\% of parameters, ensuring efficiency. Moreover, it fully retains previously learned knowledge, demonstrating robustness in continual learning.

\subsection{Low Rank Adaptation with Prompt Tuning}

\begin{figure}[t!]
    \centering
    \includegraphics[width=0.8\linewidth,height=0.5\linewidth]{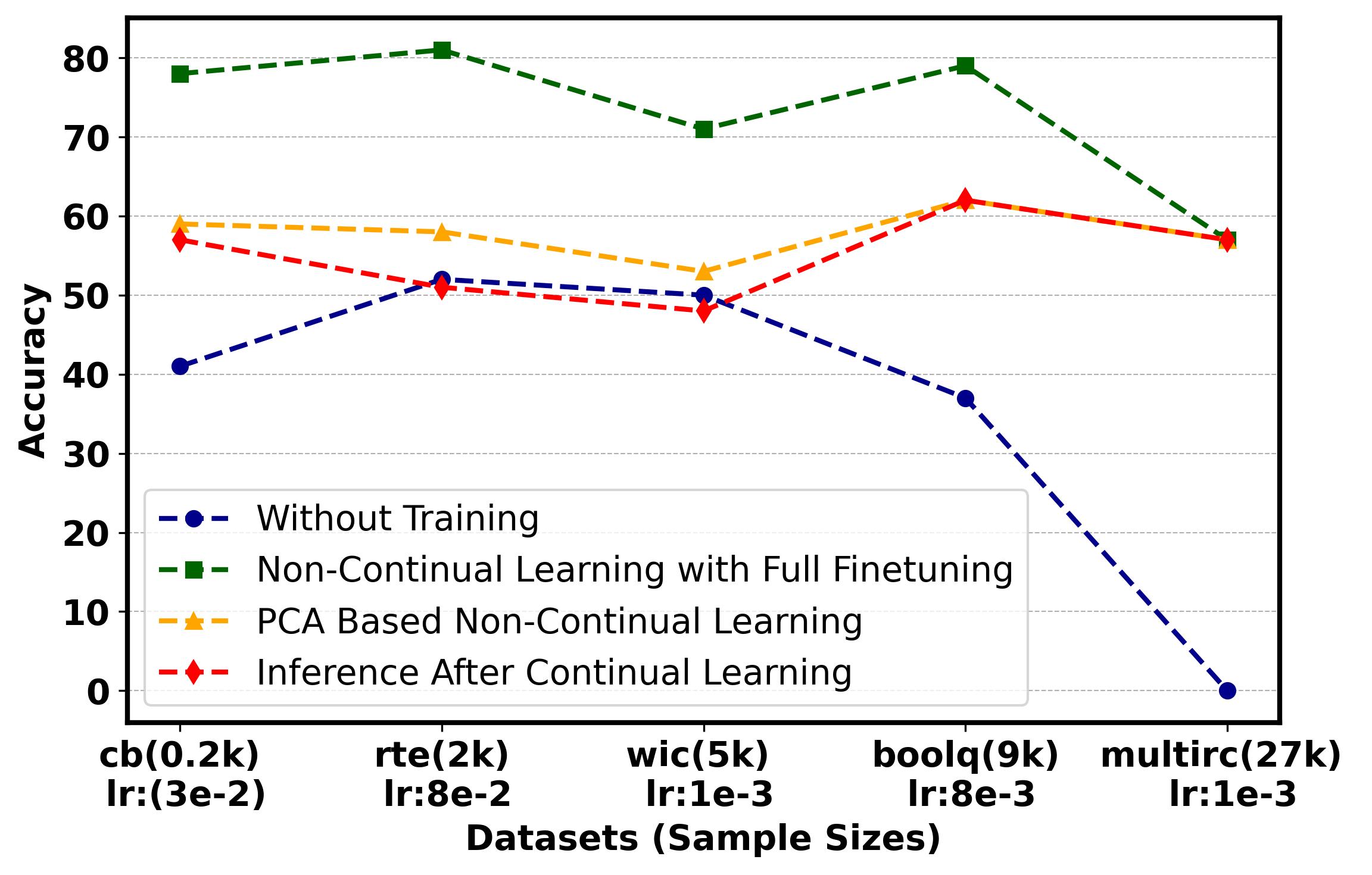}    
    \caption{\textbf{Task Incremental Learning:} Accuracy comparison of PCA-Based Continual Learning, PCA-Based Non-Continual Learning, Full Finetuning, and Zero-Shot Inference.}
\label{fig:tcl_comparison}
\end{figure}

We also evaluate a hybrid method integrating PCA-based tuning with LoRA. This tests adaptability to efficient learning strategies while enhancing accuracy without sacrificing knowledge retention. By fine-tuning only 1\% of parameters, the PCA + LoRA approach preserves all prior knowledge while improving accuracy, further reinforcing our framework’s efficiency. Fig.~\ref{fig:tcl_comparison} presents a high-level comparison of task-incremental learning results, showing that PCA-based continual learning retains prior knowledge while achieving superior accuracy compared to the zero-shot performance of the base model. This underscores the robustness of our method in adapting to new tasks without forgetting previously learned information.

Across all experiments, our framework maintains a constant training cost, regardless of task count or model size. This is achieved by limiting trainable parameters to the product of soft tokens and PCA components, typically resulting in only a few thousand parameters. For instance, with ten soft tokens and 300 PCA components, trainable parameters account for less than 0.002\% of GPT-2’s total parameters. This highlights our framework’s computational efficiency, making it well-suited for large-scale models and resource-constrained environments. Overall, our results demonstrate that the proposed framework effectively mitigates \textit{catastrophic forgetting} while achieving strong \textit{forward and backward transfer}. By combining \textbf{subspace-guided prompt reuse} with \textbf{orthogonal prompt initialization}, it ensures task-specific adaptation, knowledge retention, and scalability—addressing key challenges in continual learning for large language models.

\section{Conclusions}
We presented a subspace-guided prompt tuning framework for efficient continual learning in large language models by leveraging PCA-based subspace identification. This approach minimizes catastrophic forgetting, facilitates knowledge transfer, and significantly reduces computational overhead. Furthermore, its seamless integration with LoRA showcases its adaptability to hybrid parameter-efficient tuning strategies, achieving improved performance with a minimal number of trainable parameters.

\bibliographystyle{IEEEtran}
\bibliography{ref.bib}

\end{document}